\documentclass[10pt,twocolumn,letterpaper]{article}
\usepackage{cvpr}              
\usepackage[accsupp]{axessibility}  

\usepackage{graphicx}
\usepackage{amsmath}
\usepackage{amssymb}
\usepackage{booktabs}

\usepackage[pagebackref]{hyperref}
\hypersetup{colorlinks}

\usepackage{nicefrac}       
\usepackage{microtype}
\usepackage[dvipsnames]{xcolor}
\usepackage{multirow}

\usepackage[capitalize]{cleveref}
\crefname{section}{Sec.}{Secs.}
\Crefname{section}{Section}{Sections}
\Crefname{table}{Table}{Tables}
\crefname{table}{Tab.}{Tabs.}

\begin{document}


\title{TexPose: Neural Texture Learning for Self-Supervised \\ 6D Object Pose Estimation}

\author{%
    Hanzhi Chen$^{1}$ 
  \quad Fabian Manhardt$^{2}$   
  \quad Nassir Navab$^{1}$   
  \quad Benjamin Busam$^{1}$  
  \\ 
  $^{1}$ Technical University of Munich \quad
  $^{2}$ Google Inc.
  \\
  {{
  \tt\small hanzhi.chen@tum.de 
  \quad fabianmanhardt@google.com 
  \quad b.busam@tum.de}
  }
}
\maketitle

\begin{abstract}
In this paper, we introduce neural texture learning for 6D object pose estimation from synthetic data and a few unlabelled real images. Our major contribution is a novel learning scheme which removes the drawbacks of previous works, namely the strong dependency on co-modalities or additional refinement. These have been previously necessary to provide training signals for convergence. We formulate such a scheme as two sub-optimisation problems on texture learning and pose learning. We separately learn to predict realistic texture of objects from real image collections and learn pose estimation from pixel-perfect synthetic data. Combining these two capabilities allows then to synthesise photorealistic novel views to supervise the pose estimator with accurate geometry. To alleviate pose noise and segmentation imperfection present during the texture learning phase, we propose a surfel-based adversarial training loss together with texture regularisation from synthetic data. We demonstrate that the proposed approach significantly outperforms the recent state-of-the-art methods without ground-truth pose annotations and demonstrates substantial generalisation improvements towards unseen scenes. Remarkably, our scheme improves the adopted pose estimators substantially even when initialised with much inferior performance.

\end{abstract}

\section{Introduction}
\label{sec:intro}

For spatial interaction with objects, one needs an understanding of the translation and rotation of targets within 3D space.
Inferring these 6D object pose parameters from a single RGB image is a core task for 3D computer vision. This visually retrieved information has a wide range of applications in AR/VR~\cite{DBLP:journals/tvcg/MarchandUS16, DBLP:conf/cvpr/HampaliROL20}, autonomous driving~\cite{DBLP:conf/cvpr/ManhardtKG19,  DBLP:conf/cvpr/ZakharovKBG20,DBLP:journals/ral/GasperiniHMMNBT22}, and robotic manipulation~\cite{DBLP:conf/rss/XiangSNF18, wang2021demograsp, DBLP:conf/cvpr/LabbeCAS21}. 
Noteworthy, accuracy and runtime have both recently made a huge leap forward thanks to deep learning~\cite{DBLP:conf/iccv/KehlMTIN17, DBLP:conf/iccv/RadL17, DBLP:conf/eccv/LabbeCAS20, DBLP:conf/cvpr/PengLHZB19, DBLP:conf/eccv/ManhardtKNT18}. Unfortunately, most of these methods heavily rely on a massive amount of labelled data for supervision to learn precise models with strong generalisation capabilities~\cite{DBLP:conf/rss/XiangSNF18, DBLP:conf/cvpr/PengLHZB19, DBLP:conf/eccv/HodanSDLBMRM20, DBLP:conf/iccv/ZakharovSI19}. However, it is very labor-intensive and time consuming to generate accurate annotations for pose data~\cite{DBLP:conf/rss/XiangSNF18, DBLP:conf/wacv/HodanHOMLZ17}. Meanwhile, this process also easily suffers from labelling errors as precise annotation in 3D is highly challenging~\cite{DBLP:conf/accv/HinterstoisserLIHBKN12}. Therefore, most of the benchmarks have only few hundreds images, which does not allow proper learning of large models. In fact, methods training with such low amount of real data tend to strongly overfit to the data domain and fail to generalise to new scenes~\cite{DBLP:conf/iccvw/KaskmanZSI19}. 

As a consequence, many different approaches have been proposed in the literature to tackle this problem. The simplest solution is to employ cut-and-paste strategy to increase domain invariance~\cite{DBLP:journals/ijcv/LiWJXF20, DBLP:conf/iccv/DwibediMH17}. Nonetheless, this requires highly accurate manual annotations and most models tend to overfit to the original domain. Another alternative is to rely on a large amount of synthetically generated data to prevent overfitting. Though this process is relatively cheap and fast, rendered images can exhibit drastic visual discrepancy in comparison with real images even when advanced physically-based renderers~\cite{DBLP:journals/corr/abs-1911-01911} are used. A handful of approaches try to close the domain gap via the use of generative adversarial networks to translate the synthetic images into the real domain~\cite{DBLP:conf/cvpr/BousmalisSDEK17, DBLP:conf/iccv/ZakharovKI19}. Unfortunately, these methods do not achieve promising results as the generated images are still easily distinguishable from real imagery. Notably, very recently a new line of work that proposes to self-supervise the pose estimator on real data  has emerged. After training in simulation they fine-tune on the new datasets~\cite{DBLP:conf/eccv/WangMSJNT20, Wang_2021_self6dpp}. While these methods achieve impressive results that are even on par with fully supervised methods, they still suffer from significant performance drop when not leveraging additional supervisory signals such as depth data~\cite{Wang_2021_self6dpp} or ground truth camera poses~\cite{shugurov2021dpodv2, li2022nerf}. In this work, we propose a novel way to conduct self-supervised 6D pose estimation that is free from any additional supervision sources and further yields state-of-the-art performance.

The core idea of previous attempts~\cite{DBLP:conf/eccv/WangMSJNT20, DBLP:conf/3dim/SockGAK20, DBLP:conf/cvpr/YangY021, Wang_2021_self6dpp} to adapt pretrained pose estimators to the real domain is to conduct \textit{render-and-compare} using a differentiable renderer. With CAD models and poses given, a renderer can output several attributes of the current object pose (e.g. mask, colour, depth) which allows to refine the pose through iterative comparison between the renderings and observations. 
However, when relying on 2D visual contents (mask and colour), such strategies tend to fail due to measured silhouette imperfection, lack of textures, and domain discrepancies. This undesirable behaviour is also discussed in~\cite{DBLP:conf/eccv/WangMSJNT20}.

Thus, different from the previous attempts that heavily rely on \textit{render-and-compare} for self-supervision~\cite{DBLP:conf/eccv/WangMSJNT20, Wang_2021_self6dpp}, we instead propose to regard realistic textures of the objects as an intermediate representation before conducting training for the pose estimator.

Our approach is formulated as two interconnected sub-optimisation problems on texture learning and pose learning. In the core, we first learn realistic textures of objects from raw image collections, then synthesise training data to supervise the pose estimator with pixel-perfect labels and realistic appearance. The key challenge of our proposed scheme lies in capturing accurate texture under noise introduced by poses initialised by a pretrained pose estimator during supervision. To this end, in addition to leveraging synthetic data to establish geometry priors, we learn robust supervision through adversarial training by conditioning synthesised colours on local surface information. Furthermore, we establish regularisation from synthetic textures to compensate segmentation artefacts during a texture-learning phase. We demonstrate that the proposed approach significantly outperforms the recent state-of-the-art methods without ground-truth pose annotations and demonstrates substantial generalisation improvements towards unseen domains. Our method significantly improves even difficult objects with little variance in appearance and geometry through self-supervision. Impressively, Our approach demonstrates a robust self-improving ability for the employed pose estimators even when initialised with much inferior pose estimates than stronger baselines~\cite{Wang_2021_self6dpp}.

To summarise, our main contributions are:
\vspace{-0.2cm}
\begin{itemize}
\item[$\bullet$] We formulate a new learning scheme, \textbf{TexPose}, that decomposes self-supervision for 6D object pose into \textbf{Tex}ture learning and \textbf{Pose} learning.
\item[$\bullet$] We propose a surfel-conditioned adversarial training loss and a synthetic texture regularisation term to handle pose errors and segmentation imperfection during texture learning, further delivering self-improving ability to the pose estimators.
\item[$\bullet$] We show significant improvements over recent strong baselines with additional supervision signals. Our pose estimators demonstrates a substantial generalisation ability even on unseen scenes.
\end{itemize}
\vspace{-0.2cm}
\section{Related Work}

\paragraph{Model-based 6D Pose Estimation}
To retrieve the 6D pose, early methods use local or global features and search for key points correspondence on CAD models~\cite{DBLP:conf/eccv/BayTG06, DBLP:conf/iccv/Lowe99, DBLP:conf/icra/ColletS10, DBLP:journals/ijrr/ColletMS11}. In recent years, learning-based methods dominate the field and solve the task using convolution neural networks (CNN) under the supervision from annotated data to extract deep features. There are two major approaches for pose estimation, in particular, correspondence-based and regression/classification-based approaches. Correspondence-based methods establish 2D-3D correspondences \cite{DBLP:conf/cvpr/PengLHZB19, DBLP:conf/iccv/RadL17, DBLP:conf/iccv/ParkPV19, shugurov2021dpodv2, DBLP:conf/iccv/ZakharovSI19}, prior to leveraging a variant of the RANSAC\&PnP paradigm to solve for pose. Regression-based approaches, on the other hand, directly regress or classify the pose of the detected object. Initially these methods usually have shown a lower performance due to the existence of ambiguities~\cite{manhardt2019explaining} such as pose symmetries~\cite{DBLP:conf/rss/XiangSNF18}. Therefore, methods like SSD-6D \cite{DBLP:conf/iccv/KehlMTIN17} discretise the rotation space to circumvent this issue. Recently, with better continuous representations for rotation~\cite{DBLP:conf/cvpr/ZhouBLYL19}, these methods gradually demonstrate high effectiveness~\cite{DBLP:conf/cvpr/0001MTJ21, DBLP:conf/iccv/DiM0JNT21}. 

\paragraph{Self-Supervised Pose Learning}
Considering tremendous effort to collect large amount of annotations for 6D object pose~\cite{PhoCal,jung2022my}, several recent works have been proposed to explore the possibility of self-supervised pose learning using labeled synthetic data together with unlabelled real sensor data. \cite{DBLP:conf/icra/DengXMEBF20} designed a novel labelling pipeline using a manipulator to generate reliable pose annotations for supervision. Self6D~\cite{DBLP:conf/eccv/WangMSJNT20} proposed a self-supervision  workflow by first pretraining a pose estimator with synthetic data, which was then adapted to the real world through a \textit{render-and-compare} strategy by imposing consistencies between the rendered depth and sensed depth under the current pose estimate. CPS$++$~\cite{manhardt2020cps++} used similar approach for categorical-level object pose estimation, and the shape is jointly deformed during optimisation. AAE~\cite{sundermeyer2018implicit} parameterise $SO(3)$ space using a latent embedding obtained from synthetic images, which is later employed for orientation retrieval. Both works demonstrate the strong requirement for depth data in order to allow for a reasonable performance either in training or testing. However, due to the uninformative appearance and the sim-to-real domain gap, \textit{render-and-compare} strategies easily diverge when there is no depth data available as shown in~\cite{DBLP:conf/eccv/WangMSJNT20}. Hence, consecutive works introduce other supervisory signals to prune the need for depth. DSC-PoseNet~\cite{DBLP:conf/cvpr/YangY021} employs a key point consistency regularisation for dual-scale images with labelled 2D bounding box. Sock\etal~\cite{DBLP:conf/3dim/SockGAK20} use photometric consistency from multiple views to refine the raw estimate, while ground-truth masks are required for supervision. Recently, Self6D++~\cite{Wang_2021_self6dpp} proposed to initialise the \textit{render-and-compare} process with a powerful deep pose refiner~\cite{DBLP:journals/ijcv/LiWJXF20} to guide the learning process of the pose estimator, which currently yields state-of-the-art performance among all methods without manual annotations. Chen \etal~\cite{chen2022sim} leverages a tailored heuristics for pseudo labelling under student-teacher learning scheme.

\paragraph{Novel View Synthesis} 
With a few images given, novel view synthesis aims to render scene content from unseen viewpoints, which can be regarded as a process to acquire "textures" of the scene. NOL~\cite{DBLP:conf/eccv/ParkPV20} is proposed to address the difficulty of CAD model texturing for 6D pose estimation. Though requiring pose annotations, it demonstrates high potential to gain improvement for pose estimators with more synthesised real data in a semi-supervised fashion. Outside the scope of 6D pose, neural scene representation as implicit scene models recently get significant attention due to their ability for photorealistic novel view synthesis. NeRF~\cite{DBLP:conf/eccv/MildenhallSTBRN20} represent scenes using simple multi-layer perceptrons to encode the density and radiance for every 3D point. Assuming precise poses to transform views into a common reference frame, NeRF is able to capture volumes using an image reconstruction loss. NeRF-w~\cite{DBLP:conf/cvpr/Martin-BruallaR21} further extends NeRF to handle changing lighting and transient effects for image collections with latent embeddings. To release the rigid requirement of accurate poses, NeRF$--$~\cite{DBLP:journals/corr/abs-2102-07064} jointly optimises camera parameters and radiance fields. iNeRF~\cite{yen2020inerf} performs a \textit{render-and-compare} strategy using a trained NeRF for pose refinement. In contrast, BARF \cite{DBLP:conf/iccv/LinM0L21} demonstrates promising results when encountering imperfect pose initialisation by bundle-adjusting NeRF and camera poses with coarse-to-fine positional encoding scheme. GNeRF \cite{DBLP:conf/iccv/MengCLW0X0Y21} formulates the learning of radiance fields and coarse pose estimates as a generative modelling process and then jointly refines both using photometric information. Recently, NeRF-Supervision \cite{yen2022nerfsupervision} and NeRF-Pose \cite{li2022nerf} are proposed to leverage NeRF to learn dense descriptors or 3D representations of a given object under pose annotations when CAD model is inaccessible.
\section{Method}
Our objective is to conduct per-instance 6D pose learning from monocular colour images with no manual labelling effort, namely \textit{self-supervised 6D pose estimation} originally defined in~\cite{DBLP:conf/eccv/WangMSJNT20}. Since the CAD model is known as a prior, we follow the two-stage approach of previous works \cite{DBLP:conf/eccv/WangMSJNT20, DBLP:conf/3dim/SockGAK20, DBLP:conf/cvpr/YangY021,yen2020inerf, Wang_2021_self6dpp}, which first pretrain a pose estimator with labelled synthetic data, and consequently conduct self-training with unannotated real images captured from a specific real scene demanding a more precise deployment (e.g. robotic bin picking). We mainly introduce the self-training process in this section. Notably, in this tuning process, pose estimator tends to suffer from forgetting problem due to little variance in illumination and backgrounds of the fed training samples. In contrast, We will show the strength of our scheme to alleviate this issue in Section.~\ref{sec:exp}. After pretraining we acquire a pose estimator $f_{\theta}$ able to initialise a coarse pose estimate $T$ and a segmentation mask $M$ for each real image $\hat{I}$. $M$ is usually used as a pseudo silhouette cue to refine the pose $T$ with \textit{render-and-compare} strategy \cite{DBLP:conf/cvpr/KunduLR18, DBLP:conf/eccv/ChenD0GH20, DBLP:conf/eccv/WangMSJNT20, yen2020inerf}. 

In this section, we first revisit the commonly used strategy of previous methods in this task, then introduce our new formulation requiring no additional supervisory sources clearly advantageous over previous attempts. 

\subsection{Revisiting Self-Supervised 6D Pose Learning}
\begin{figure*}[!t]
      \centering
      \includegraphics[width=0.9 \textwidth,clip, trim=0.1cm 0.1cm 0.1cm 0.1cm]{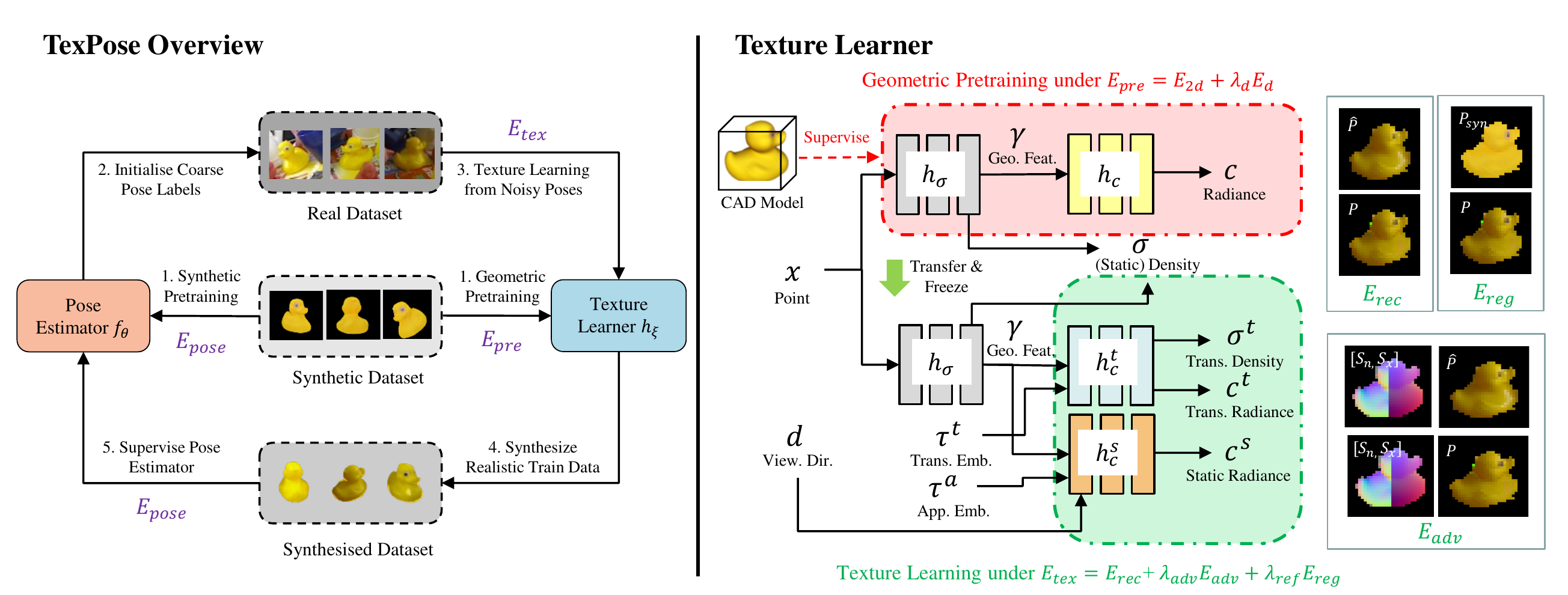}
 
      \caption{Left: Schematic overview of TexPose. We fully explore the benefit of both synthetic and real dataset with perfect geometric labels and realistic appearance to synthesise informative train dataset for pose learning. Right: Details of the texture learner used in our scheme. We first conduct geometric pretraining to acquire geometry branch $h_{\sigma}$ to ensure consistent coordinates system and geometry aligned with the CAD model. Then, we transfer and freeze $h_{\sigma}$ to the real together with newly established radiance branch $h_{c}^{s}$ and $h_{c}^{t}$ as our texture learner. It is supervised by $E_{rec}$, $E_{reg}$ and $E_{adv}$ capable of improving texture quality so as to synthesise more informative dataset for supervision.}
      \label{fig:pipeline}
\end{figure*}

\textit{Render-and-compare} is a common approach for 6D pose self-supervision~\cite{DBLP:conf/eccv/WangMSJNT20, DBLP:conf/3dim/SockGAK20, DBLP:conf/cvpr/YangY021,yen2020inerf, Wang_2021_self6dpp}. As shown in Eqn.~\ref{eqn:poseopt}, it utilises a differentiable renderer $\mathcal{R}$ fed with an estimated pose to output a rendered mask, image, $[M_r, I_r]  = \mathcal{R}(T)$. $E_{rend}$ enforces consistency between the rendered outputs $[M_r, I_r]$ and the observations $[M, \hat{I}]$. It aims to refine the pseudo pose labels for each training sample, and further supervise pose estimator $f_{\theta}$ under $E_{pose}$. Previous attempts are majorly focused on improving $E_{rend}$ to acquire more accurate pseudo pose annotations for better supervision. 
\begin{equation}
\begin{aligned}
\min_{\theta, \{ {T}_{i}, \ldots, {T}_{N} \} } \quad &  
                    \sum_{i}  E_{rend}(\mathcal{R}(T_i), [M_i, \hat{I}_i])    \\+ & \sum_{i} E_{pose}(f_{\theta}(\hat{I}_i), T_i). 
\end{aligned}
\label{eqn:poseopt}
\end{equation}
When relying on 2D information $[M_r, I_r]$, minimising $E_{rend}$ to improve $\{ {T}_{1}, \ldots, {T}_{N} \}$ can be extremely difficult due to uninformative shape and appearance and sim-to-real visual discrepancy (e.g., Duck in Fig.~\ref{fig:pipeline}). To compensate for the lack of information, related methods introduce additional supervision with varying difficulty levels for the acquisition (e.g, depth \cite{DBLP:conf/eccv/WangMSJNT20}, weak labels \cite{DBLP:conf/3dim/SockGAK20, DBLP:conf/cvpr/YangY021}, deep pose refiner \cite{Wang_2021_self6dpp}, etc.). Deep pose refiner could be acquired with the least effort as it only requires synthetic pretraining as well. However, it can introduce a strong upper bound as illustrated in \cite{Wang_2021_self6dpp}. The pose refiner can even lead to divergence of the pose estimator and hence lacks robustness for geometrically and photometrically challenging objects. 
 
\subsection{TexPose: Learning Pose from Texture}
 
To address the aforementioned limitations, we reformulate self-supervised 6D pose learning as two alternating optimisation problems for a texture learner $h_{\xi}$ and a pose estimator $f_{\theta}$ as defined in Eqn.~\ref{eqn:texopt} \footnote{\textcolor{ForestGreen}{Green} indicates weight update and \textcolor{RedOrange}{orange} indicates weight freezing.} and depicted in Fig.~\ref{fig:pipeline} (Left). 

\begin{equation}
\begin{aligned}
\min_{\theta, \xi} \quad &  \sum_{i} E_{tex}(\textcolor{ForestGreen}{h_{\xi}} \circ \textcolor{RedOrange}{f_{\theta}}(\hat{I}_i), \hat{I}_i)  \\+ & \sum_{j} E_{pose}(\textcolor{ForestGreen}{f_{\theta}} \circ \textcolor{RedOrange}{h_{\xi}}(\hat{T_j}), \hat{T_j}). 
\end{aligned}
\label{eqn:texopt}
\end{equation}

Instead of leveraging raw images as regression targets for poses, we insert object textures as an intermediate representation before pose learning. Specifically, we start with optimising $E_{tex}$ by fixing the parameters of pose estimator \textcolor{RedOrange}{$f_{\theta}$} used to initialise poses  $\{ {T}_{1}, \ldots, {T}_{N} \}$ for each raw image $\{ \hat{I}_{1}, \ldots, \hat{I}_{N} \}$, aiming to provide supervision for the realistic textures embedded by \textcolor{ForestGreen}{$h_{\xi}$}. Afterwards, we freeze \textcolor{RedOrange}{$h_{\xi}$} to synthesise training samples $\{ {I}_{1}, \ldots, {I}_{N^\prime} \}$ paired with perfect labels $\{ \hat{T}_{1}, \ldots, \hat{T}_{N^\prime} \}$. Gradients will thus be propagated through \textcolor{ForestGreen}{$f_{\theta}$} to update its parameters during the minimisation for $E_{pose}$ over these train data. We typically chose $N^\prime > N$.  

With this reformulation, we deliberately omit pose optimisation from 2D information usually sensitive to initialisation and prone to divergence, and bridge the domain gap directly through photorealistic synthesis. Our goal is then shifted to limit texture mapping error under the presence of pose noise so that the synthesised data can be highly informative with perfect geometric labels and as-accurate-as-possible realistic appearance. Intuitively, through multi-view supervision, it's possible to capture more accurate appearance with compensation for per-view pose noise. Our proposed strategy to further increase reliability of the synthesised data will be discussed 
later. Notably, different from previous attempts, no additional supervision is required in our design. The pose estimator is equipped with a self-improving ability through intermediate phase for texture learning.   
 
\paragraph{NeRF preliminaries.} 
We leverage neural radiance fields (NeRF) \cite{DBLP:conf/eccv/MildenhallSTBRN20} to embed the texture information due to its simplicity and capability of photorealistic view synthesis~\footnote{Radiance, texture and appearance are discussed as interchangeable concepts in our paper.}. 

 NeRF represents the object model with two MLPs to encode information for every queried point $\mathbf{x}$, namely the geometry branch to predict volume density $\sigma$ and geometric feature $\mathbf{\gamma}$ with $[\sigma, \mathbf{\gamma}] = h_{\sigma}(\mathbf{x})$ and radiance branch for colour $\mathbf{c}$ prediction with $ \mathbf{c} = h_{c}(\mathbf{x}, \mathbf{\gamma})$ (weight $\xi$ is omitted). When rendering from a given viewpoint, $K$ samples $\{ x_k \}_1^K$ are gathered along the ray $\mathbf{r}=\mathbf{o} + t\mathbf{d}$ where $\mathbf{o}, \mathbf{d} \in \mathbb{R}^3$ represent the camera centre and the view direction respectively and $t \in \mathbb{R}$ is the distance. The colour $\mathbf{C}(\mathbf{r})$ will be estimated through numerical quadrature approximation with $w_{k} = \exp(- \delta_k \sigma_k)$ and $\alpha_{k} = \prod_{j=1}^{k-1} w_{j}$ with $\delta_k$ being the step size between adjacent points, and $\mathbf{C}(\mathbf{r}) = \sum_{k=1}^{K} \alpha_{k}(1 - w_{k})\mathbf{c_k}$. Rendering for depth $D(\mathbf{r})$ and mask $M(\mathbf{r})$ is achieved by modifying integration element (see supp. mat. for details).
\paragraph{Geometric pretraining.}
Though geometry and radiance branch can be trained jointly as shown in NeRF and its variants \cite{DBLP:conf/eccv/MildenhallSTBRN20,  DBLP:conf/cvpr/Martin-BruallaR21, neroic}, the pose for supervision in our scenario is highly noisy, and the coordinate system of NeRF can end up as an arbitrary one and further cause erroneous labels for synthesised data. We hence opt to pretrain geometry branch \textcolor{ForestGreen}{$h_{\sigma}$} using pixel-perfect data rendered by the CAD model with accurate pose $\hat{T}$ , depth $\hat{D}$ and mask $\hat{M}$ to supervise NeRF following Eqn.~\ref{eqn:pretrain}, where $E_d$ is scale-invariant loss from~\cite{kopf2021robust}, $\lambda_{m}$ and $\lambda_{d}$ are weighting factors.

\begin{equation}
\begin{aligned}
E_{2d}(\mathbf{r}) = \left \Vert \mathbf{C}(\mathbf{r}) - \mathbf{\hat{C}}(\mathbf{r})\right \Vert_{2}^{2}  +  \lambda_{m} \left \Vert M(\mathbf{r}) - \hat{M}(\mathbf{r})\right \Vert_{2}^{2} 
\end{aligned}
\label{eqn:pretrain2d}
\end{equation}

\begin{equation}
\begin{aligned}
E_{pre}(\mathbf{r}) = E_{2d}(\mathbf{r}) + 
    \lambda_{d} E_{d}(D(\mathbf{r}),  \hat{D}(\mathbf{r})).
\end{aligned}
\label{eqn:pretrain}
\end{equation}
Notably, though our current goal is to establish a geometry branch with precise density output "aligned" with the CAD model, training for radiance branch is also included, though it will be discarded afterwards. We find this greatly helps to extract meaningful geometric features fed into the radiance branch, and further eases the effort for realistic appearance transfer in the later texture learning stage. We quantitatively detail the necessity of this stage in supp. mat.

\paragraph{Texture learning from noisy poses.}
 
Our current goal is to learn texture encoded by the radiance branch \textcolor{ForestGreen}{$h_{c}$} using raw images paired with poses initialised by the pose estimator \textcolor{RedOrange}{$f_{\theta}$}. Note in this stage, we only re-optimise parameters from \textcolor{ForestGreen}{$h_{c}$} with \textcolor{RedOrange}{$h_{\sigma}$} being fixed to capture accurate texture and meanwhile ensure no shift of coordinate system and geometry. Though jointly optimising textures and poses through photometric consistency has a potential to improve both aspects as shown in ~\cite{yen2020inerf, DBLP:conf/iccv/LinM0L21, neroic}, we find it can easily fail when dealing with the textureless objects. Alternatively, we seek to supervise the texture learner with higher pose error tolerance leveraging stronger radiance branch and robust loss. 

Different from the vanilla NeRF used in geometric pretraining, we now employ two radiance branches, \textcolor{ForestGreen}{$h_{c}^{s}$} and \textcolor{ForestGreen}{$h_{c}^{t}$}, to translate the extracted geometric features into static radiance (accurate underlying textures) and transient radiance (texture outliers caused by pose errors) respectively. Apart from  geometric features $\mathbf{\gamma}$, we further condition it on view directions $\mathbf{d}$, lighting embedding \textcolor{ForestGreen}{$\mathbf{\tau}^{a}$} and transient embedding \textcolor{ForestGreen}{$\mathbf{\tau}^{t}$}~\cite{DBLP:conf/cvpr/Martin-BruallaR21} (see Fig.~\ref{fig:pipeline} Right). For each view, we use \textcolor{ForestGreen}{$\mathbf{\tau}^{a}$} to capture illumination variations and \textcolor{ForestGreen}{$\mathbf{\tau}^{t}$} to produce transient effect caused by pose noise. Patch-based training strategy~\cite{DBLP:conf/nips/SchwarzLN020} is used to preserve the spatial structures for robust loss supervision. An image reconstruction loss $E_{rec}$ is incorporated with uncertainty awareness to learn textures robustly. We refer to the supp. mat. for the details of the radiance branches and uncertainty-aware $E_{rec}$.  
 
We noticed that supervision from $E_{rec}$ is still not adequate to capture accurate texture under pose noise. We further design a surfel-conditioned adversarial loss for sampled patches to tolerate texture shifts caused by pose errors. For each location on the object surface, we compute normalised object coordinate $S_x$ \cite{DBLP:conf/cvpr/Wang0HVSG19} and normal $S_n$ with current pose estimate $T$ to represent its local information. By conditioning the predicted patch colour $P$ on its local surface information $[S_x, S_n]$, we employ an adversarial scheme:
\begin{equation}
\begin{aligned}
E_{adv}(P) =  &\mathbb{E}_{[S_{x}, S_{n}], P}(1 - \log D([S_{x}, S_{n}], P))  \\  + &\mathbb{E}_{[S_{x}, S_{n}], \hat{P}}(\log D([S_{x}, S_{n}], M\hat{P})).
\end{aligned}
\label{eqn:adv}
\end{equation}
We also observed the imperfect predicted mask $M$ can cause background colour artefacts within the object. We therefore establish a regularisation from synthetic cues to compensate the boundary imperfections by generating synthetic patch $P_{syn}$ and mask $M_{syn}$ with current pose estimate $T$, then applying erosion on the predicted mask $M$ to compute a padded mask with $M_{pad} = M_{syn}(1 - M)$. The foreground boundaries of the real images for supervision are therefore padded with the synthetic colours. This regularisation $E_{feat}$ is realised by minimising the MSE error of deep features extracted by a pretrained VGG19 network~\cite{DBLP:journals/corr/SimonyanZ14a} to align the boundary features of the predicted patch $P$ with the synthetic patch $P_{syn}$ in a more abstract level. The second part of Eqn.~\ref{eqn:feat} ensures that rendered patch region within the mask $M$ are not affected by the synthetic information.
 
\begin{equation}
\begin{aligned}
E_{reg}(P) = & E_{feat}(P, M\hat{P} + M_{pad}P_{syn}) \\+ 
&\lambda_{fg} E_{feat}(MP + (1 - M)\hat{P}, \hat{P}).
\end{aligned}
\label{eqn:feat}
\end{equation}
Formally, the loss function $E_{tex}$ for the texture learner supervision is summarised as: 
\begin{equation}
\begin{aligned}
E_{tex} = E_{rec} + \lambda_{adv} E_{adv} + \lambda_{reg} E_{reg}.
\end{aligned}
\label{eqn:fulltex}
\end{equation}
where $\lambda_{adv}$ and $\lambda_{reg}$ are weighting factors.
\paragraph{Pose learning.}
The entire pipeline runs by first optimising $E_{tex}$, and subsequent minimisation of $E_{pose}$ with the synthesised dataset as depicted in Fig.~\ref{fig:pipeline} (Left). As our scheme operates at data level and thus method-agnostic, $E_{pose}$ can be generalised across methods with different loss designs, e.g., direct pose or correspondence regression. Interestingly, we observed that the pose estimator can converge instantly with only one optimisation loop on $E_{tex}$ and $E_{pose}$, while adding more loops, i.e., continue to optimise $E_{tex}$ with improved poses after $E_{pose}$, only brings insignificant improvements. We attribute this to the pixel-perfect supervision from texture learner and its strong ability to mitigate pose errors.

\begin{figure*}[!t]
      \centering
      \includegraphics[width=0.8\textwidth,clip, trim=0.3cm 0.3cm 0.3cm 0.3cm]{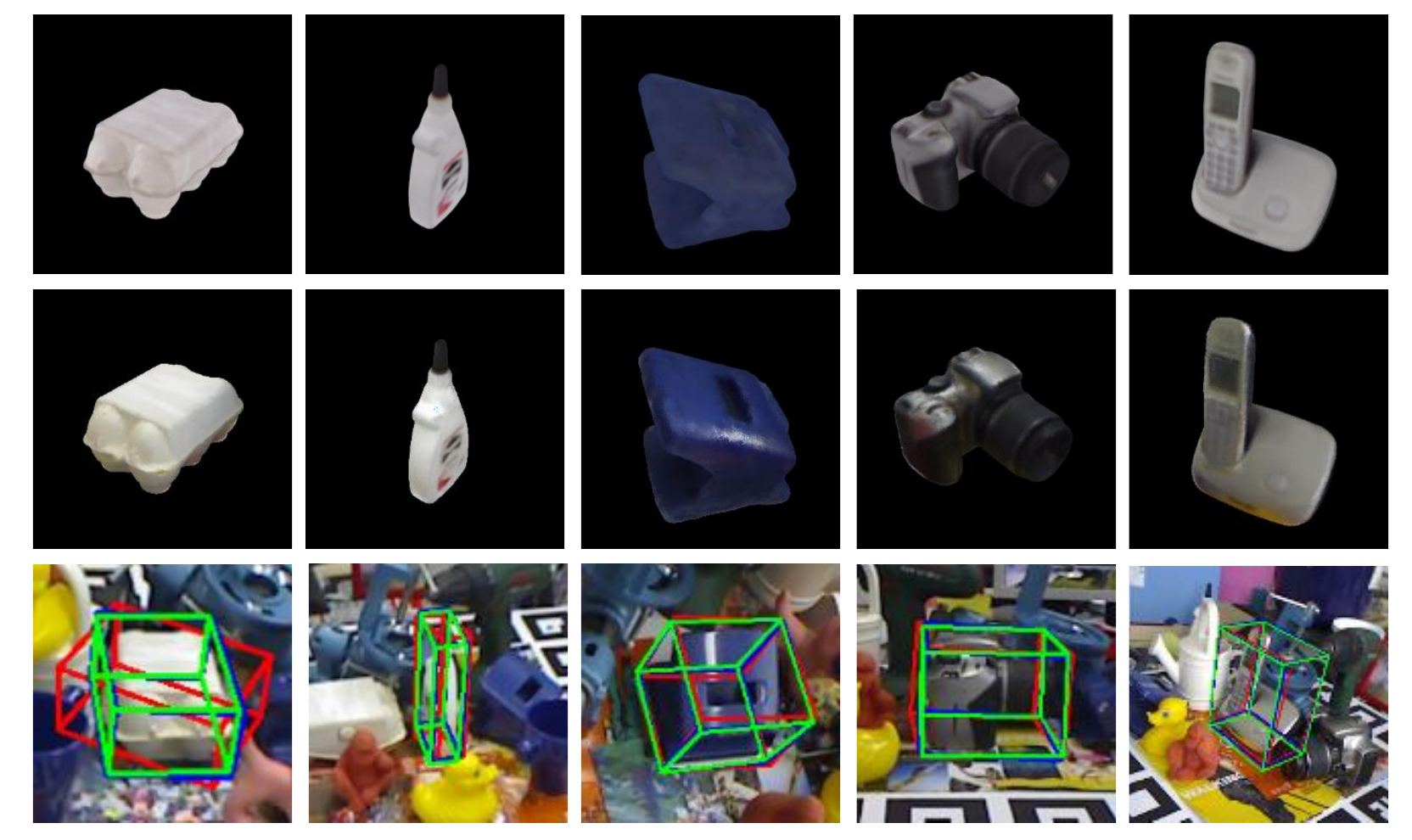}
      \caption{Qualitative results on 5 objects from LineMOD. Top: Renderings of the reconstructed CAD models which are part of LineMOD. Middle: Same objects rendered with our learnt textures. Bottom: Demonstration of our pose estimation quality. Thereby, the green, red, and blue boxes depict ground-truth pose as well as estimated pose before and after our self-supervision, respectively.  }
      \label{qualitativ}
      \vspace*{-5mm}
\end{figure*}

\section{Experiments}
In this section we first introduce the employed metrics and dataset before presenting our results for the task of 6D pose estimation, then provide detailed analysis of our proposed scheme in ablation study. 
\label{sec:exp}

\subsection{Evaluation Metrics} 
In line with previous work, we present our results using the ADD(-S) metric, as proposed in~\cite{DBLP:conf/accv/HinterstoisserLIHBKN12}. This metric reports 6D pose error by transforming all object vertices with estimated pose and ground-truth pose and measuring the average distance of the two sets of points. If the average distance is below $10\%$ of the object diameter, the estimated pose will be considered correct. For symmetric objects, ADD score is modified to measure the distance to the nearest vertices, here referred to as ADD-S. 

\subsection{Dataset}
\paragraph{Synthetic BOP PBR.} We pretrain our models using synthetic data from \cite{DBLP:conf/eccv/HodanSDLBMRM20, DBLP:journals/corr/abs-1911-01911}, employing physically-based rendering (PBR) techniques. Although PBR can capture more realistic lighting conditions, there still exhibits a large sim-to-real domain gap. Such discrepancies can strongly degrade the pose estimation performance when being deployed in the real world.

\paragraph{LineMOD.}~\cite{DBLP:conf/accv/HinterstoisserLIHBKN12} consists of 13 objects with watertight CAD models. Thereby, each object is accompanied by a small image sequence of around 1k frames, captured in cluttered environment with challenging lighting. Following previous works \cite{DBLP:conf/iccv/RadL17, DBLP:conf/eccv/WangMSJNT20}, we employ 15\% of the captured RGB images for self-supervision, discarding any manual annotations, and test our model on the remaining samples. 

\paragraph{Occluded LineMOD.} \cite{DBLP:conf/eccv/BrachmannKMGSR14} is an extension of one LineMOD sequence showing eight different objects from LineMOD undergoing severe occlusion. For fairness we adopt the same train/test split for self-supervision as \cite{Wang_2021_self6dpp}.

\paragraph{HomebrewedDB.} \cite{DBLP:conf/iccvw/KaskmanZSI19} is a recently established dataset with various kinds of objects and poses. To test generalisability we test on the three objects that it shares with LineMOD~\cite{DBLP:conf/iccvw/KaskmanZSI19}. 

\setlength{\tabcolsep}{3.8pt}
\begin{table*}[!t]
\begin{center}
\caption{Evaluation results on LineMOD dataset. *: objects with symmetry. **: used as our pretrained pose estimator before self-supervision, consistent with Self6D++. $\dagger$: re-implemented version from \cite{Wang_2021_self6dpp}, used as supervision source for Self6D++. $\ddagger$: variant with depth supervision}
\label{table:lm}
\scriptsize
\begin{tabular}{|c | c c c c c | c c  c | c c  c c c c|}

\hline
Supervision & \multicolumn{5}{|c|}{Syn} & \multicolumn{3}{c|}{Syn + Real} & \multicolumn{6}{c|}{Syn + Self} \\

\hline
\multicolumn{15}{|c|}{Supervision signals required in real domain}\\
\hline
 Depth      & -  & - & - & - & - & -   & - & - & \checkmark & - & - & - &  \checkmark  & -\\
 
 Pose Refiner & -  & - & - & - & - & -   & - & - & - & - & - & \checkmark & \checkmark & -\\
 
 Weak labels & -  & - & - & - & - & -   & - & - & - & \checkmark & \checkmark & - & - & -\\
 
 GT Pose labels & -  & - & - & - & - & \checkmark & \checkmark & \checkmark & - & - & - & - & - & -\\
  
 
 \hline \hline


\multirow{2}{*}{Methods}  & AAE & MHP & DPODv2  & GDR-LB** &  DeepIM$\dagger$  & DPODv2 & GDR-UB &                             SO-Pose & Self6D & Sock~\etal & DSC & Self6D++ & Self6D++$\ddagger$ &                              \multirow{2}{*}{ \textbf{Ours}}
            \\

                      & 
                      \cite{sundermeyer2018implicit} & 
                      \cite{manhardt2019explaining}&
                      \cite{shugurov2021dpodv2}& 
                      \cite{DBLP:conf/cvpr/0001MTJ21} &  \cite{DBLP:journals/ijcv/LiWJXF20} & 
                      \cite{shugurov2021dpodv2} & 
                      \cite{DBLP:conf/cvpr/0001MTJ21} & \cite{DBLP:conf/iccv/DiM0JNT21} & \cite{DBLP:conf/eccv/WangMSJNT20} & 
                      \cite{DBLP:conf/3dim/SockGAK20} & 
                      \cite{DBLP:conf/cvpr/YangY021} & 
                      \cite{Wang_2021_self6dpp} & 
                      \cite{Wang_2021_self6dpp} & \\

\hline
Ape       & 4.0  & 11.9 & 62.1 & 50.9 & \textbf{85.8} & 80.0 & \textbf{85.0} &- & 38.9 & 37.6 & 31.2 & 76.0 & 75.4  & \textbf{80.9}\\
Benchvise & 20.9 & 66.2 & 88.3 & \textbf{99.4} & 93.1 & 99.7 & \textbf{99.8} & - & 75.2 & 78.6 & 83.0 & 91.6 & 94.9 & \textbf{99}\\
Camera    & 30.5 & 22.4 & 92.5 & 89.2 & \textbf{99.1} & \textbf{99.2} & 96.5 &- &  36.9 & 65.6 & 49.6 & \textbf{97.1} & 97.0 & 94.8\\
Can       & 35.9 & 59.8 & 96.6 & 97.2 & \textbf{99.8} & \textbf{99.6} & 99.3 &- &  65.6 & 65.6 & 56.5 & \textbf{99.8} & 99.5 & 99.7\\
Cat       & 17.9 & 26.9 & 86.1 & 79.9 & \textbf{98.7} & \textbf{95.1} & 93.0 &- &  57.9 & 52.5 & 57.9 & 85.6 & 86.6 & \textbf{92.6}\\
Driller   & 24.0 & 44.6 & 90.1 & 98.7 & \textbf{100.0} & \textbf{98.9} & 100.0 &- &  67.0 & 48.8 & 73.7 & 98.8 & \textbf{98.9} & 97.4\\
Duck      & 4.9  & 8.3  & 54.8 & 24.6 & \textbf{61.9} & \textbf{79.5} & 65.3 &- &  19.6 & 35.1 & 31.3 & 56.5 & 68.3 & \textbf{83.4}\\
Eggbox*    & 81.0 & 55.7 & 98.6 & 81.1 & \textbf{93.5} &  99.6 & \textbf{99.9} &- &  99.0 & 89.2 & 96.0 & 91.0 & \textbf{99.0} & 94.9\\
Glue*      & 45.5 & 54.6 & 95.4 & 81.2 & \textbf{93.3} & \textbf{99.8} & 98.1 &- &  94.1 & 64.5 & 63.4 & 92.2 & \textbf{96.1} & 93.4\\
Holep.    & 17.6 & 15.5 & 27.0 & \textbf{41.9} & 32.1 & 72.3 & \textbf{73.4} &- &  16.2 & 41.5 & 38.8 & 35.4 & 41.9 & \textbf{79.3}\\
Iron      & 32.0 & 60.8 & 98.2 & 98.8 & \textbf{100.0} & \textbf{99.4} & 86.9 &- &  77.9 & 80.9 & 61.9 & 99.5 & 99.4 & \textbf{99.8}\\
Lamp      & 60.5 &  -   & 91.0 & 98.9 & \textbf{99.1}  & 96.3 & \textbf{99.6} &- &  68.2 & 70.7 & 64.7 & 97.4 & \textbf{98.9} & 98.3\\
Phone     & 33.8 & 34.4 & 74.3 & 64.3 & \textbf{94.8} & \textbf{96.8} & 86.3  & -&  50.1 & 60.5 & 54.4 & 91.8 & \textbf{94.3} & 78.9\\
\hline
Average   & 31.4 & 38.8 & 81.2 & 77.4 & \textbf{88.0} & 93.5 & 91.0 & \textbf{96.0}  &  58.9 & 60.6 & 58.6 & 85.6 & 88.5 & \textbf{91.7}\\ 
\hline

\end{tabular}
\end{center}
\vspace*{-3mm}
\end{table*}

\subsection{Performance Analysis}
\paragraph{Performance on LineMOD and Occluded LineMOD}

In Table~\ref{table:lm}, we demonstrate our results for LineMOD with respect to the ADD(-S) metric. Note that GDR-LB and GDR-UB correspond to the performance of our base pose estimators, respectively supervised with synthetic data and real data, reflecting the lower and upper performance bound. As one can easily see there is a significant performance gap between fully-supervised methods and approaches trained with synthetic data instead (\textit{c.f.} DPODv2~\cite{shugurov2021dpodv2}, GDR~\cite{DBLP:conf/cvpr/0001MTJ21}). Self-supervised approaches have proven to be able to almost completely close the domain gap, yet typically require additional modalities to achieve comparable results, as for example Self6D++ \cite{Wang_2021_self6dpp} with depth supervision. In contrast, our method is able to further close the domain gap whilst relying solely on RGB images, outperforming all other self-supervised approaches by at least 3\% for RGB-D approaches and 6\% for RGB-only methods. Noticeably, we are able to improve our lower bound performance from 77.4\% to 91.7\%, which is even slightly better than the performance of our upper bound model trained with full supervision. 

We would like to emphasise that our model is capable of providing significant performance boosts on highly challenging tiny and textureless objects (e.g., Ape, Duck, and Holep.) by improving the lower bound from $50.9\%, 24.6\%, 32.1\%$ to $80.9\%, 83.4\%, 79.3\%$, respectively. These objects are typically hard to handle by other self-supervised methods without real supervision (\textit{c.f.} results from Self6D++ \cite{Wang_2021_self6dpp}, Sock \etal \cite{DBLP:conf/3dim/SockGAK20}, etc). As the RGB variant of Self6D++~\cite{Wang_2021_self6dpp} leverages DeepIM to refine the output pose in order to generate pseudo pose annotations, this process imposes a strong \textit{upper bound} on the pose estimator. In fact, the performance achieved by Self6D++ is clearly inferior to DeepIM for almost all objects. This problem is even more apparent when the pose refiner is worse than the pose estimator (\textit{c.f.} Holep. results from GDR~\cite{DBLP:conf/cvpr/0001MTJ21}, DeepIM~\cite{DBLP:journals/ijcv/LiWJXF20} and Self6D++~\cite{Wang_2021_self6dpp}). On the contrary, our method relies on texture optimisation to synthesise realistic training samples with precise pose annotations for supervision, making self-supervision more robust towards pose errors. We further validate our scheme with another pose estimator, DPODv2~\cite{shugurov2021dpodv2} in supp. mat.
As for our results on Occluded LineMOD in Table~\ref{table:lmo_partial}, we can make the same observations. We again report the best overall performance of all self-supervised methods, despite not relying on any elaborate design for occlusion-awareness as Self6D++\cite{Wang_2021_self6dpp}, with 66.7\% vs 64.7\% and 59.8\% for Self6D++\cite{Wang_2021_self6dpp} with and without depth data, respectively.

\begin{figure*}[!t]
      \centering
      \includegraphics[width=0.95\textwidth,clip, trim=0.0cm 0.0cm 0.0cm 0.0cm]{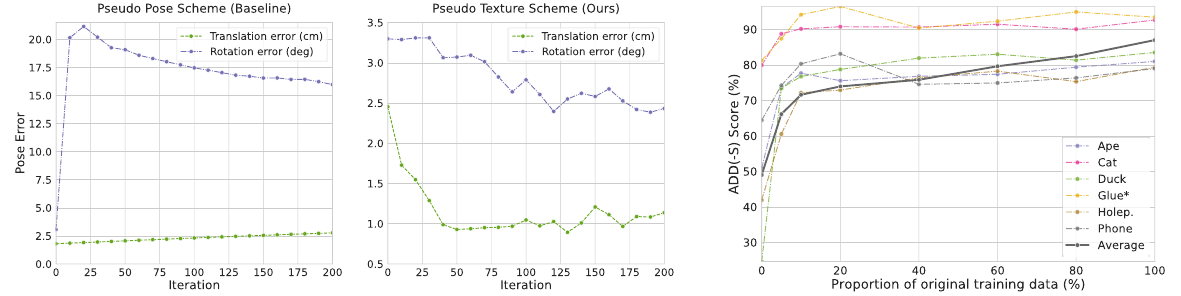}
      \caption{Ablation of the different schemes and the required amount of data for self-supervision. Left: Pose errors after every 10 iterations for the baseline scheme (pseudo pose) and our proposed pseudo-texture scheme. Right: Self-supervision with different amount of training data.} 
      \label{fig:ratio_tex}
      \vspace*{-5mm}
\end{figure*}

\setlength{\tabcolsep}{3.0 pt}
\begin{table}[!t]
\begin{center}
\caption{Evaluation results on Occluded LineMOD dataset. *: objects with symmetry. **: used as our pretrained pose estimator before self-supervision, consistent with Self6D++. $\dagger$:re-implemented version from \cite{Wang_2021_self6dpp}, used as supervision source for Self6D++. $\ddagger$: variant with depth supervision.}
\label{table:lmo_partial}
\scriptsize
\scalebox{0.95}{
\begin{tabular}{|c | c | c c c c c c|}
 
 \hline

Supervision & \multicolumn{1}{c|}{Syn} 
&\multicolumn{6}{c|}{Syn + Self} \\
\hline

\multirow{2}{*}{Methods}  & 
GDR**   & 
Self6D  & 
Sock~\etal  & 
DSC   & 
Self6D++   & 
Self6D++$\ddagger$   & 
\multirow{2}{*}{\textbf{Ours}} \\

  & 
\cite{DBLP:conf/cvpr/0001MTJ21} & 
\cite{DBLP:conf/eccv/WangMSJNT20}& 
\cite{DBLP:conf/3dim/SockGAK20} & 
\cite{DBLP:conf/cvpr/YangY021} & 
\cite{Wang_2021_self6dpp} & 
\cite{Wang_2021_self6dpp} & 
\\
\hline
Ape      & 44.0   & 13.7 & 12.0 & 9.1 & 57.7 & 59.4 & \textbf{60.5} \\
Can       & 83.9  & 43.2 & 27.5 & 21.1 & 95.0 & \textbf{96.5} & 93.4 \\
Cat       & 49.1 & 18.7 & 12.0 & 26.0 & 52.6 & \textbf{60.8} & 56.1 \\
Driller   & 88.5 & 32.5 & 20.5 & 33.5 & 90.5 & 92.0 & \textbf{92.5} \\
Duck      & 15.0  & 14.4 & 23.0 & 12.2 & 26.7 & 30.6 & \textbf{55.5} \\
Eggbox*    & 33.9  & 57.8 & 25.1 & 39.4 & 45.0 & \textbf{51.1} & 46.0 \\
Glue*      & 75.0  & 54.3 & 27.0 & 37.0 & 87.1 & \textbf{88.6} & 82.8 \\
Holep.    & 34.0 & 22.0 & 35.0 & 20.4 & 23.5 & 38.5 & \textbf{46.5} \\
\hline
Average   & 52.9 & 32.1 & 22.8 & 24.8 & 59.8 & 64.7 & \textbf{66.7} \\ 
\hline

\end{tabular}
}
\end{center}
\vspace{-5mm}
\end{table}
\setlength{\tabcolsep}{2.3pt}
\begin{table}[htb]
\begin{center}
\caption{Evaluation results on HomebrewedDB dataset. **: used as our pretrained pose estimator before self-supervision, consistent with Self6D++. $\dagger$: Pre-process testing images with the same intrinsics as LineMOD, different from \cite{Wang_2021_self6dpp} as they directly feed raw images without scaling. $\ddagger$: Use raw images for re-adaptation.}
\label{table:hb_partial}
\scriptsize
\begin{tabular}{|c | c | c c c c | }
\hline
Supervision & Syn & \multicolumn{4}{c|}{Syn + Self}  \\
\hline  
Method    
& GDR** $\dagger$ \cite{DBLP:conf/cvpr/0001MTJ21}   
& Sock~\etal  \cite{DBLP:conf/3dim/SockGAK20}
& Self6D++ $\dagger$ \cite{Wang_2021_self6dpp}
& \textbf{Ours} $\dagger$   
& \textbf{Ours} $\ddagger$ \\

\hline
Benchvise & 88.8   & 57.3  & 75.7  & \textbf{93.1} & 92.9 \\
Driller   & 92.8   & 46.6  & 89.4  & \textbf{94.8} & 94.2 \\
Phone     & 78.7   & 41.5  & 76.8  & 79.3 & \textbf{81.2}\\
\hline
Average   & 86.8   & 52.0  & 80.6  & \textbf{89.1} & 89.4\\
\hline

\end{tabular}
\end{center}
\vspace{-5mm}
\end{table}

\vspace*{-3mm}
\paragraph{Performance on HomebrewedDB and generalisation}

In this experiment, we want to study the generalisation ability of our method towards new scenes with changing illumination. 

From Table~\ref{table:hb_partial}, we observe that after self-supervision on LineMOD, Self6D++ undergoes a clear performance drop compared to their respective pretrained models when testing on HomebrewedDB (86.8\% vs. 80.6\%). This reveals that Self6D++'s self-supervision scheme is prone to forgetting. Our method can instead generalise to the new scene with even boosted performance (86.8\% vs. 89.1\%). We attribute this to the learnt realistic textures, which are used to uniformly synthesise photorealistic data that can further be augmented with random backgrounds to increase domain invariance~\cite{DBLP:journals/ijcv/LiWJXF20, DBLP:conf/iccv/DwibediMH17}. Further, we investigate the adaptation ability of our method on these new scenes. To this end, we conduct self-supervision on raw images from HomebrewedDB. We first synthesise novels view with the pre-trained texture learner from LineMOD in an effort to adapt the pose estimator to the camera from HomebrewedDB. Afterwards, we re-train the texture learner with raw images from the new target scene. Our results again illustrate that our method outperforms all other baselines. Interestingly, we notice that the choice of adaptation domain (LineMOD or HomebrewedDB) has marginal effect on our final performance, suggesting that our method has potential to close the sim-to-real domain gap even trained for one scene.

\subsection{Ablation Study}
 
\paragraph{Dealing with bad initialisation.} We randomly sample 200 testing images from the Duck sequence of LineMOD (shown in Fig.~\ref{fig:pipeline}). For the baseline \textit{render-and-compare} scheme (pseudo pose), we minimise $E_{rend}$ in Eqn.~\ref{eqn:poseopt} to refine the pose with the raw initialisation from the pose estimator. For our proposed scheme (pseudo texture), we repeatedly run an evaluation after every 10 weight update steps of the pose estimator. From Fig.~\ref{fig:ratio_tex} (left), we see that the baseline scheme diverges, while our proposed scheme significantly enhances the convergence properties thanks to the synthesised supervision from the texture learner. 

\setlength{\tabcolsep}{2.3pt}


\begin{table}[!h]
\begin{center}

\caption{Ablation study on partial objects from LineMOD dataset. *: objects with symmetry.} 

\centering
\scalebox{0.92}{
\begin{tabular}{| c | c | c c c c c c | c|}
\hline
              $E_{adv}$ & $E_{reg}$ & Ape & Cat & Duck & Glue* & Holep. & Phone & Average \\
\hline
\multicolumn{2}{|c|}{w/o self-sup.} & 50.9          & 79.9          &           24.6 &          81.2 & 41.9 & 64.3 & 57.1   \\
\hline

  - & -                     & 76.3          & 90.3          &           65.9 &          82.3 & 59.6 & 65.3 & 73.3   \\
  \checkmark & -            & 80.4          & 89.4          &           76.5 &          90.9 & 76.5 & 70.7 & 80.7   \\
  - & \checkmark            & \textbf{82.1} & 87.8          &           78.4 & \textbf{96.4} & 77.1 & 76.2 & 83.0   \\
  \checkmark & \checkmark   & 80.9          & \textbf{92.6} & \textbf{83.4}  &          93.4 & \textbf{79.3} & \textbf{78.9} & \textbf{84.7}   \\

\hline

\end{tabular}}
\label{table:ablative}

\end{center}
\end{table}
\vspace*{-5mm}
\paragraph{Impact of $E_{adv}$ and $E_{reg}$.} 
We study the individual components of our proposed training objective to guide texture learning under pose noise using the six worst performing objects. As shown in Table~\ref{table:ablative}, even when training the texture learner with noisy poses, we are able to significantly improve the lower bound for most objects. We attribute this behaviour to the strong synthesis capabilities of NeRF~\cite{DBLP:conf/eccv/MildenhallSTBRN20}.  
Notably, for objects with richer textures or articulation (Glue, Phone), the baseline model without their supervision yields nearly no improvements. However, when respectively introducing $E_{adv}$ and $E_{reg}$ to the baseline model, we observe ca. $10\%$ absolute improvement on ADD(-S) score. Further, combining both of them yields the best performance.
\raggedbottom
\paragraph{Impact of training data size.} As the core idea of our method is to learn the object texture from multiple views, instead of optimising per-sample pose for supervision, we believe such formulation has a potential to allow for pose learning with much less data. To this end, we conduct self-supervision with an increasing dataset size taken from the original training split. On the right of Fig.~\ref{fig:ratio_tex}, we can see that the performance keeps steadily increasing when the amount of data is enlarged. Surprisingly, we observe that already with 5\% of the training split (only 9 images), our method can bring a significant absolute improvement of ca. $18\%$, achieving around $75\%$ of the optimal performance. This further confirms our method can even serve as a strong baseline for few-shot learning under extreme data scarcity. 

We refer interested readers to supp. mat for further detailed ablation experiments, e.g., analysis on optimisation design (Eqn.~\ref{eqn:poseopt} and Eqn.~\ref{eqn:texopt}) in Table~\textcolor{red}{7}. 

\section{Conclusion}
\label{sec:conclusion}
In this paper, we formulate a new learning paradigm for the task of 6D pose estimation. Separating the task into texture and pose learning enables a novel self-supervision scheme to adapt pretrained 6D pose estimators to the real domain. Only a few unlabelled images are needed within this framework to generate pixel-perfect supervision. With this reformulation, we successfully bridge the gap between synthetic pre-training and real data usage through geometry-transfer from pixel-perfect in silico data onto data-induced appearance information. It successfully addresses the difficulty in optimising pose with raw 2D observations and further prunes the need for additional supervisions as guidance. Although initialised with much inferior accuracy, it improves the state-of-the-art self-supervision RGB method by an impressive 6.1\% while even being superior to RGBD methods by 3.2\%. We further demonstrates substantial generalisation improvements towards real-world scenarios. 

For limitations, though we require as little as nine images for texture learning, they shall have mild occlusions to provide sufficient appearance information of the objects. We also share current limitations with other model-based pose estimation methods and hence need a CAD model to ensure a precise object-centric neural representation. For future works, we strongly believe that our reformulation can impact self-supervision pipelines also for unseen objects with joint optimisation of both geometry and texture.


{\small
\bibliographystyle{ieee_fullname}
\bibliography{literature}
}
\end{document}